\documentclass[letterpaper, 10 pt, conference]{ieeeconf}  

\IEEEoverridecommandlockouts                              

\usepackage{graphics} 
\usepackage{epsfig} 
\usepackage{times} 
\usepackage{amsmath} 
\usepackage{amssymb}  
\usepackage{graphicx} 

\usepackage{url}
\usepackage{cite}
\usepackage{enumerate}
\usepackage{booktabs}
\usepackage{tabularx}
\usepackage{multirow}
\usepackage[table,xcdraw]{xcolor}
\usepackage{xspace}
\usepackage{float} 

\usepackage[pagebackref=true,breaklinks=true,letterpaper=true,colorlinks,bookmarks=false]{hyperref}

\newcolumntype{x}[1]{>{\centering\arraybackslash}p{#1pt}}
\newcolumntype{y}[1]{>{\raggedright\arraybackslash}p{#1pt}}
\newcolumntype{z}[1]{>{\raggedleft\arraybackslash}p{#1pt}}

\definecolor{baselinecolor}{gray}{.9}
\newcommand{\baseline}[1]{\cellcolor{baselinecolor}{#1}}
\newcommand{\tablestyle}[2]{\setlength{\tabcolsep}{#1}\renewcommand{\arraystretch}{#2}\centering\footnotesize}

\renewcommand{\paragraph}[1]{\vspace{1.25mm}\noindent\textbf{#1}}

\title{\LARGE \bf
HAMF: A Hybrid Attention-Mamba Framework for Joint Scene Context Understanding and Future Motion Representation Learning
}

\author{Xiaodong Mei, Sheng Wang, Jie Cheng, Yingbing Chen and Dan Xu
\thanks{Xiaodong Mei, Sheng Wang, Jie Cheng, Yingbing Chen and Dan Xu are with the Hong Kong University of Science and Technology, Hong Kong SAR, China. \texttt{{xmeiab, swangei, jchengai, ychengz}@connect.ust.hk, danxu@cse.ust.hk}. \textit{(Corresponding author: Dan Xu.)}}
}

\begin{document}

\maketitle
\thispagestyle{empty}
\pagestyle{empty}

\begin{abstract}

Motion forecasting represents a critical challenge in autonomous driving systems, requiring accurate prediction of surrounding agents' future trajectories.
While existing approaches predict future motion states with the extracted scene context feature from historical agent trajectories and road layouts, they suffer from the information degradation during the scene feature encoding. 
To address the limitation, we propose HAMF, a novel motion forecasting framework that learns future motion representations with the scene context encoding jointly, to coherently combine the scene understanding and future motion state prediction.
We first embed the observed agent states and map information into 1D token sequences, together with the target multi-modal future motion features as a set of learnable tokens. 
Then we design a unified Attention-based encoder, which synergistically combines self-attention and cross-attention mechanisms to model the scene context information and aggregate future motion features jointly.
Complementing the encoder, we implement the Mamba module in the decoding stage to further preserve the consistency and correlations among the learned future motion representations, to generate the accurate and diverse final trajectories.
Extensive experiments on Argoverse 2 benchmark demonstrate that our hybrid Attention-Mamba model achieves state-of-the-art motion forecasting performance with the simple and lightweight architecture. 

\end{abstract}


\section{INTRODUCTION}

\begin{figure}[t]
\begin{center}
\includegraphics[width=0.99\linewidth,trim=0 250 190 0,clip]{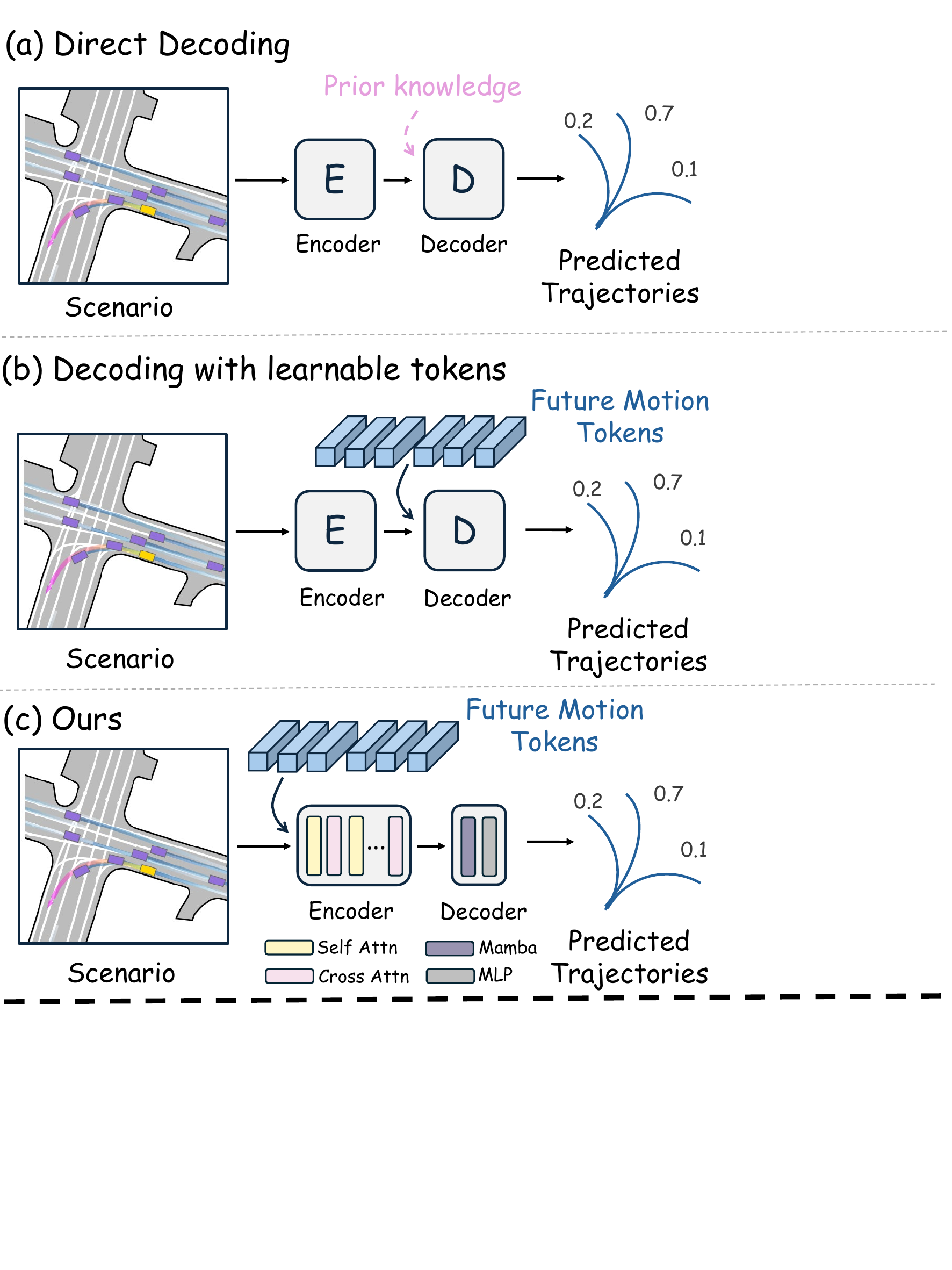} 
\end{center}
\vspace{-1em}
   \caption{Comparison of existing motion forecasting frameworks (a, b) and our proposed method (c). 
   The bounding boxes in purple denote the surrounding traffic participants and the box in yellow denotes the focal agent. 
   Historical trajectories are presented with the gradient blue lines and the ground-truth future trajectory as the prediction target is presented with the gradient pink line.   
   Differently from direct-decoding approaches and learnable anchor-based methods, we formulate the future motion feature as a set of learnable tokens and input them into the encoder with embedded scene context tokens, to acquire more comprehensive future motion representations within the scene understanding stage.}
\label{fig:intro}
\vspace{-1.2em}
\end{figure}

Motion forecasting constitutes a critical component in autonomous driving systems~\cite{wilson2023argoverse}, empowering ego vehicles to comprehensively understand dynamic surroundings and execute safe and efficient navigation strategies through probabilistic prediction of other traffic participants' future states.
The inherent multi-modality of agent behaviors, coupled with the complex spatio-temporal interactions between dynamic agents and static road layouts, presents persistent challenges in motion forecasting, particularly in accurately modeling and reliably generating diverse future behavioral modalities.
%

Existing learning-based motion forecasting frameworks typically adopt an encoder-decoder architecture~\cite{huang2022survey}, where the encoder module captures heterogeneous scene context information for the scene understanding, followed by the decoder module that generates multi-modal future trajectories conditioned on the learned feature representations.
According to the future trajectory generation mechanism, current methods can be roughly classified into two categories: direct-decoding methods~\cite{gao2020vectornet, liang2020learning, zhang2024simpl, ngiam2021scene, cheng2023forecast, cui2023gorela},  and learnable anchor-based methods~\cite{varadarajan2022multipath++,shi2022motion,zhou2023query, zhang2024decoupling, wang2023prophnet}. 
The direct-decoding method predicts future trajectories by decoding the same agent feature into various moving modes.
Despite the modeling simplicity,
these methods are prone to encountering the mode collapse issue~\cite{chai2019multipath} by producing a limited set of frequent motion patterns.
To overcome the shortcoming, several methods integrate prior knowledge into the decoding stage to improve multi-modal trajectory prediction through the incorporation of constrained trajectory hypotheses, such as surrounding lanes~\cite{deo2022multimodal}, dense goal candidates~\cite{ zhao2021tnt, gu2021densetnt} and possible trajectory proposals~\cite{song2021learning, chai2019multipath}.
However, the quality of hand-crafted priors influences the prediction performance severely.
In contrast, learnable anchor-based approaches demonstrate enhanced flexibility in capturing diverse future motion modalities, such as MTR~\cite{shi2022motion} and QCNet~\cite{zhou2023query}.
Future motion features are formulated as the learnable embeddings with or without the guidance of prior knowledge that serve as dynamic queries, enabling their adaptive combination with scene features for multi-modal trajectory generation.
%

Nevertheless, it is crucial to note the current framework limits the future motion representation learning to the decoding stage, where they are generated from the learned and hand-crafted features or formulated as supplementary learnable tokens that interact with encoded scene context features utilizing sophisticated architectural designs.
This limitation raises an intriguing question: \textit{Could the incorporation of future motion feature learning, potentially embedded as learnable tokens from the very beginning stage of scene encoding, yield more comprehensive and physically plausible representations?}
%

A fundamental advantage of incorporating future motion tokens into the encoding stage is that it facilitates the motion pattern learning across multi-level scene context representations within each encoder layer, rather than being constrained to potentially oversimplified features from the final encoding layer.
The establishment of long-range interactions between scene context features and future motion features within the encoder enables the acquisition of more comprehensive motion representations.
Some works in neural language processing~\cite{devlin2018bert} and computer vision~\cite{alexey2020image} have demonstrated the feasibility of the idea, where they directly input one learnable classification token within the sequence into the self-attention based encoder module to obtain the category representation. 
Furthermore, TaskPrompter~\cite{taskprompter2023} inputs a set of learnable tokens that guides the multi-task image feature learning through the combination of self-attention and cross-attention modules.
We take it a step further to propose HAMF, a hybrid Attention-Mamba model presented in Figure \ref{fig:intro} that directly incorporates the multi-modal future motion feature learning within the scene encoding stage, as a novel and unified framework to combine the scene context understanding and future motion state prediction. 
We represent the future motion feature as a sequence of learnable tokens, which are combined with embedded observed agent tokens and map tokens to form the input of the encoder module. 
The unified encoder employs a sequential stack of self-attention and cross-attention blocks~\cite{vaswani2017attention} in each layer, which interactively and progressively extract and refine future motion feature representations.
Besides, due to the strong sequence modeling ability of the recent state space model Mamba~\cite{gu2023mamba}, we further design a simple yet effective Mamba-based decoder to model the relationship of the multiple future motion tokens, to ensure accuracy and diversity. Our contributions are as follows:

\begin{itemize}

\item We propose a novel motion forecasting framework HAMF that unifies the scene context understanding and future motion state prediction.  By coherently incorporating multi-modal future motion representation learning with scene encoding, our model is capable of aggregating more comprehensive and feasible future trajectory features. 
\item We propose a hybrid model architecture comprising an Attention-based encoder and a Mamba-based decoder. 
The encoder employs a sequential stack of self-attention and cross-attention blocks within each layer, enabling comprehensive extraction of global scene context and effective modeling of future motion representations.
Then the Mamba-based decoder captures the complex dependencies among multiple future motion modalities, to generate both diverse and accurate trajectories. 
\item Our approach achieves competitive performance with state-of-the-art methods on the challenge Argoverse 2 benchmark with the simpler and more lightweight model architecture. 

\end{itemize}

\section{RELATED WORK}
\subsection{Learning-based Motion Forecasting}
Motion forecasting has witnessed significant advancements in recent years, benefiting from comprehensive open-source datasets~\cite{ wilson2023argoverse} and innovations in deep learning.
This essential module involves cross-modal scene contextual feature extraction and future state prediction of target agents.
The scene representation has progressed from early rasterized aerial view images~\cite{tang2019multiple,chai2019multipath} to current vectorized paradigms~\cite{gao2020vectornet} enabled by high-definition (HD) mapping technologies. 
Several works~\cite{liang2020learning, chen2022hgcn, jia2023hdgt} then utilize graph-based representations to jointly model the agent motion information and detailed road elements.
The integration of Transformer architectures~\cite{ngiam2021scene, zhang2024simpl, zhang2024decoupling, zhang2024real, zhou2023query} has further advanced the field through tokenized scene representations and Attention-based feature processing to enhance the prediction capability significantly.
Additionally, various forecasting frameworks integrate the prior knowledge to guide the prediction, including predefined trajectory proposals~\cite{song2021learning, chai2019multipath}, possible goals~\cite{zhao2021tnt, gu2021densetnt} or anchor points~\cite{shi2022motion}. 
Furthermore, self-supervised pre-training methodology~\cite{cheng2023forecast} and adaptive post-refinement strategy~\cite{zhou2024smartrefine} are proposed to improve the prediction performance. 

In contrast to existing approaches, we propose a novel framework that directly integrates multi-modal future motion feature learning into the scene encoding phase, to establish a new paradigm that unifies scene understanding and future state prediction.

\subsection{Attention and Mamba}
Attention mechanism based models~\cite{vaswani2017attention} have demonstrated remarkable advancements across diverse domains\cite{alexey2020image, devlin2018bert, carion2020end,cheng2024rethinking,cheng2024pluto}, primarily through two fundamental paradigms: self-attention and cross-attention.
The self-attention mechanism captures intra-sequence contextual dependencies, whereas cross-attention facilitates inter-sequence relationship modeling between source and target sequences, thereby enhancing output coherence. 
Despite the achievements, the quadratic computational complexity remains an inherent limitation in attention mechanisms, necessitating further architectural innovations.
The development of state space models (SSMs) has introduced a compelling alternative in sequence modeling with recently proposed Mamba architecture~\cite{gu2023mamba}.
This selective state space model combines the expressive power of Transformer-based approaches with linear computational scaling, enabling significant performance improvements in various applications ranging from computer vision~\cite{liu2025vmamba, li2024videomamba} to multi-modal representation learning~\cite{zhao2024cobra}.

Leveraging the model architectural developments, our approach strategically combines the complementary strengths of Mamba's efficient sequence modeling and attention-based feature representation learning, yielding impressive performance improvements.

\begin{figure*}[hbt!]
\begin{center}
\vspace{0.8em}
\includegraphics[width=0.93\linewidth, trim= 5 0 85 5]{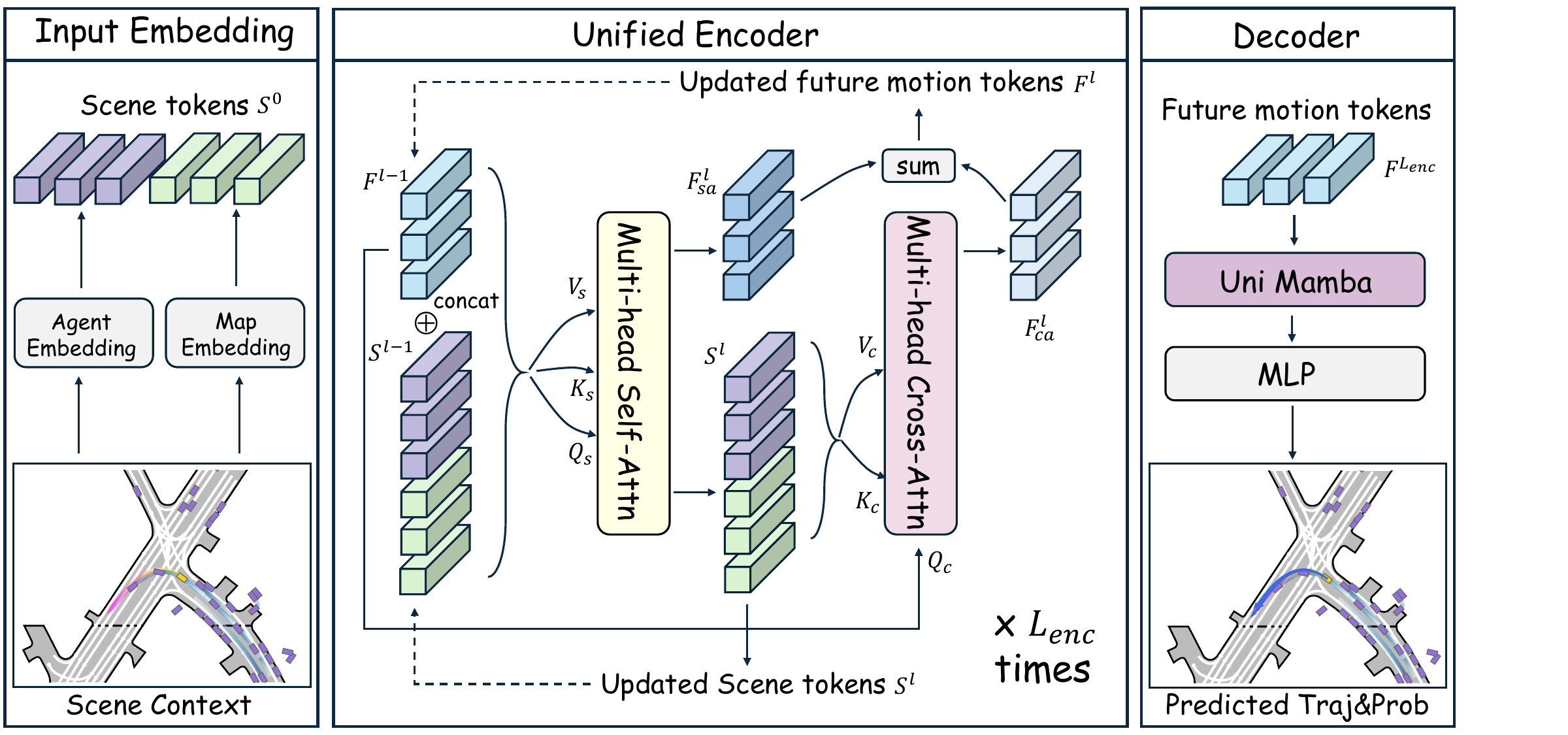}
\vspace{-1em}
\end{center}
   \caption{
   Overview of our proposed HAMF.
   The left part presents the input embedding module with an intersection driving scenario. The ground truth future trajectory is shown with the gradient pink line for illustration purposes, which is not used in the input. The historical trajectories and surrounding map are embedded and combined as initial scene tokens $S^0$, then concatenated with the initial future motion tokens $F^0$ for the input of the unified encoder. 
   The middle part denotes the encoding process within the $l$-th encoder layer. The dashed lines represent the input for the subsequent encoding layer.
   With $L_{enc}$ iterations in the encoder, the learned future motion tokens $F^{L_{enc}}$ are obtained and decoded with the Mamba block and multi-layer MLPs to generate the final prediction, shown in the right part. 
   }
\label{fig:model}
\vspace{-1em}
\end{figure*}

\section{METHODS}

In this section, we present our motion forecasting model HAMF with the Attention-based encoder and Mamba-based decoder. 
The overall framework is presented in Figure \ref{fig:model}. 

\subsection{Problem Formulation} 
Given a driving scenario with the HD map, we follow the popular vectorized representation to organize $N$ surrounding agents' trajectories and $M$ lane segments as polylines.
Specifically, the historical trajectories of agents are represented as $A_{in} \in \mathbb{R}^{N_{in} \times T_h \times C_a}$, where $N_{in}$ is the number of input agents, $T_h$ is the number of historical time steps and $C_a$ is the dimension of input motion state (e.g., step-wise displacement/velocity difference, observation status of certain time step). 
Similarly, the map is denoted as $M \in \mathbb{R}^{M \times L \times C_m}$, where $M$ is the number of map polylines, $L$ is the number of points of each polylines and $C_m$ is the feature channels. 
With the observed agent states and map information, the task of motion forecasting is to generate $K$ multi-modal future trajectories of the target agent, denoted as $P \in \mathbb{R}^{ K \times T_f \times 2}$ with $T_f$ indicating the number of future time steps. The probability score of each future trajectory is required to be predicted simultaneously.
We employ the same data preprocessing and normalization strategy as in our previous work~\cite{cheng2023forecast}.

\subsection{Input Representation and Embedding}
We first utilize individual embedding modules to encode the agent and map features into 1D tokens. 
Like ~\cite{shi2022motion, cheng2023forecast}, we employ the PointNet-based embedding model to generate map tokens $X_m \in \mathbb{R}^{M \times C}$, where $C$ indicates the embedding dimension.
For agent tokens $X_a \in \mathbb{R}^{N_{in}} \times C$, we follow~\cite{zhang2024decoupling} to employ the stacked unidirectional Mamba~\cite{mamba} blocks to capture the motion correlations and dependency during the observed time steps. We concatenate agent and map tokens to form the complete scene tokens $S \in \mathbb{R}^{(N_{in} + M) \times C}$.
The scene embedding process is formulated as
\begin{gather}
X_a = \text{UniMamba}(A_{in}), 
\\
X_m = \text{MiniPointNet}(M), 
\\
S = \text{Concat}(X_a, X_m),
\end{gather}
where the agent category and lane type information, together with the position embedding (PE), are added to scene tokens as the learnable embeddings.

In order to incorporate the future motion features into the scene encoding stage, we formulate $K_e$ multi-modal future trajectory features into 1D learnable tokens as $F \in \mathbb{R}^{K_e \times C}$ without any hand-crafted priors. 
Note that $K_e$ could be different from the required number $K$ of future predictions, as each motion token represents one kind of motion modality. 

\subsection{Unified Encoder}
The goal of our unified encoder is to model the scene context features and aggregate the future motion features concurrently. 
Previous works successfully employ the self-attention mechanism to encode scene context from the combined agent and map tokens, while other works employ the cross-attention mechanism to capture the future trajectory features with the motion query.
Therefore, we introduce the Attention-based encoder module that synergistically integrates self-attention and cross-attention mechanisms. 
The module incorporates both attention blocks within each encoder layer sequentially and interactively, to leverage their complementary strengths.

Specifically, every encoder layer contains a self-attention block followed by a cross-attention block.
We take the $l$-th encoder layer to explain details, where $l \in \{1, 2,...,L_{enc}\}$ denotes the number of layers. For illustration purposes, we omit the standard residual connection, layer normalization and feed-forward process in multi-head attention blocks.

For scene tokens ${S}^{l-1}$ and future motion tokens $F^{l-1}$ obtained from the previous encoder layer, we first concatenate them to form the token sequence $X^{l-1} \in \mathbb{R}^{(K_e+N_{in} + M ) \times C}$, which is the complete input of the self-attention block.
\begin{gather}
X^{l-1} = \text{Concat}(F^{l-1}, {S}^{l-1}), 
\\
X^{l} = \text{MHA}(Q_s=X^{l-1}, K_s=X^{l-1}, V_s=X^{l-1}).   
\end{gather}

We add the Layer Normalization layer~\cite{ba2016layer} to the output of the self-attention block and split it to obtain the updated future motion tokens $F^{l}_{sa}$ and the current scene context features $S^{l}$. 
\begin{gather}
F^{l}_{sa}, S^{l} = \text{Split}(X^{l}).   
\end{gather}

Then we perform the cross-attention mechanism to further capture the future motion features from the scene context features in the current layer.
\begin{gather}
F^{l}_{ca} = \text{MHA}(Q_c=F^{l-1}, K_c=S^{l}, V_c=S^{l}).
\end{gather}

Finally, to interactively combine and leverage the strength of both attention algorithms, we simply add the updated future motion tokens from the self-attention block and the cross-attention block without introducing more parameters into the model to get $F^{l}$,
\begin{gather}
F^{l} = F^{l}_{sa} + F^{l}_{ca}.
\end{gather}

The updated future motion tokens together with the updated scene tokens $S^{l}$ constitute the input of the next encoder layer for the same learning procedure. Specially, the initial tokens $S^0$ and $F^0$ come from the respective embedding modules directly.

\subsection{Multi-modal Decoder}
With the feature aggregation procedure in the encoder, the future motion tokens are subsequently used to predict the multi-modal future trajectories for the target agent by the one-token-to-one-trajectory paradigm.
To better cover the possibilities of diverse motion patterns, we utilize the unidirectional Mamba module to model the dependency among the series of $K_e$ future motion tokens, due to its effectiveness and efficiency in the sequential modeling.
\begin{gather}
F' = \text{UniMamba}(F^{L_{enc}}).
\end{gather}

Then final trajectories and the corresponding probabilities are generated with multiple layers of MLP respectively. 
To facilitate the model to generate more scene-compliant trajectories, we also adopt another separate multi-layer MLP as the prediction head to predict one future trajectory for each surrounding agent.

\subsection{Training Loss}
We typically use the winner-take-all strategy to employ the trajectory regression loss and the classification loss for probability without loss weights as in~\cite{zhang2024decoupling}.

\begin{table*}[t]
\vspace{0.8em}
\caption{Comparison results on Argoverse 2 test set. For all metrics, the lower is the better. \textbf{The upper group} presents results with single model and \textbf{the lower group} presents results with model ensembling strategy. We \textbf{bold} the best results and \underline{underline} the second best results.}
\centering
\def\arraystretch{1.1} 
\vspace{-0.8em}
\resizebox{0.95\textwidth}{!}{
\begin{tabular}{x{80}x{40}x{40}x{40}x{40}x{40}x{50}}
\toprule
 Method & minADE$_1 \downarrow$ & minFDE$_1 \downarrow$ & minADE$_6 \downarrow$ & minFDE$_6 \downarrow$ & MR$_6 \downarrow$ & b-minFDE$_6 \downarrow$ \\ 
 \midrule
 HDGT~\cite{jia2023hdgt}  & 2.08 & 5.37  & 0.84& 1.60 & 0.21 & 2.24 \\
 SIMPL~\cite{zhang2024simpl} & 2.03  & 5.50 & 0.72  & 1.43 & 0.19 & 2.05 \\
 GoRela~\cite{cui2023gorela}& 1.82  & 4.62 & 0.76 & 1.48 & 0.22 & 2.01 \\
 ProphNet~\cite{wang2023prophnet}  & 1.80 & 4.74 & 0.68 & 1.33 & 0.18 & \underline{1.88} \\
 GANet~\cite{wang2023ganet}  & 1.77 & 4.48 & 0.72 & 1.34 & 0.17 & 1.96 \\
MTR~\cite{shi2022motion}  & 1.74 & 4.39 & 0.73 & 1.44 & \underline{0.15} & 1.98 \\
Forecast-MAE~\cite{cheng2023forecast} & 1.74  & 4.36  & 0.71 & 1.39 & 0.17 & 2.03 \\
 QCNet~\cite{zhou2023query}  & 1.69 & 4.30 & 0.65 & 1.29 & 0.16 & 1.91 \\
SmartRefine~\cite{zhou2024smartrefine}  & 1.65 & 4.17 & \textbf{0.63} & 1.23 & \underline{0.15} & \textbf{1.86} \\
 DeMo~\cite{zhang2024decoupling}  & \underline{1.60} & 4.00 & 0.65 & 1.25 & \underline{0.15} & 1.92 \\
RealMotion~\cite{song2024realmotion}  & 1.59 & \textbf{3.93} & 0.66 & \underline{1.24}& \underline{0.15} & 1.89 \\

 \textbf{HAMF (Ours)}  & \baseline{\textbf{1.59}} & \baseline{\underline{3.99}}  & \baseline{\underline{0.64}} & \baseline{\textbf{1.23}} & \baseline{\textbf{0.14}} & \baseline{1.89} \\
 \hline 
 QML~\cite{su2022qml} & 1.84 & 4.98  & 0.69 & 1.39 & 0.19 & 1.95\\
 MacFormer~\cite{feng2023macformer}  & 1.84 & 4.69  & 0.70 & 1.38 & 0.19 & 1.90\\ 
  BANet~\cite{zhang2022banet}  & 1.79 & 4.61 & 0.71 & 1.36  & 0.19 & 1.92 \\
 Gnet~\cite{gao2023dynamic}  & 1.72 & 4.40  & 0.69 & 1.34 & 0.18 & 1.90 \\
 Forecast-MAE~\cite{cheng2023forecast}  & \underline{1.66} & 4.15 & 0.69 & 1.34  & \underline{0.17} & 1.91 \\
 QCNet~\cite{zhou2023query}  & 1.56 & \underline{3.96} & \textbf{0.62} & \textbf{1.19}  & 0.14 & \textbf{1.78} \\
\textbf{HAMF (Ours)}  & \baseline{\textbf{1.56}} & \baseline{\textbf{3.91}}  & \baseline{\underline{0.63}} & \baseline{\underline{1.20}} & \baseline{\textbf{0.14}} & \baseline{\underline{1.81}} \\
\bottomrule
\end{tabular}
}
\label{tab:leaderboard}
\vspace{-0.8em}
\end{table*}


\section{EXPERIMENTS}

\subsection{Experimental Setup}

\subsubsection{Dataset}
We use the large-scale real-world motion forecasting dataset Argoverse 2 (AV2)~\cite{wilson2023argoverse} to evaluate the proposed HAMF. 
The dataset contains 250K interesting and challenging scenarios collected from six geographically diverse cities, with each lasting 11 seconds and sampled at 10 Hz. Given the provided HD map and 5 seconds historical motion states for every scenario, the task is to predict the future trajectory in 6 seconds horizon.

\subsubsection{Metrics} 
We utilize official metrics in the leaderboard: minADE$_K$, minFDE$_K$, MR$_K$ and b-minFDE$_K$. In AV2, $K$ is set to 6 for multi-modal prediction evaluation and 1 for uni-modal evaluation.

\subsubsection{Implementation Details}
Our model is trained in an end-to-end manner by 4 NVIDIA GeForce RTX 3090 GPUs for 120 epochs with around 10 hours. The batch size is 32 per GPU.
The initial learning rate is set to 0.001 with a cosine learning rate schedule.
The AdamW~\cite{loshchilov2017decoupled} optimizer and the weight decay as 0.01 are used for training.
We utilize five encoder layers and one Uni-Mamba block with two-layer MLPs in the decoder.
With the latent feature dimension set to 128, our hybrid Attention-Mamba model is quite lightweight with 3.0M parameters in total.

\subsection{Main Results}
\paragraph{Comparison with state-of-the-art.}
Our proposed HAMF (without model ensemble techniques) surpasses most 
existing approaches on the Argoverse 2 test set and demonstrates competitive results compared to state-of-the-art methods, as presented in the upper group of Table \ref{tab:leaderboard}. 
Notably, our approach achieves the best results in terms of minADE$_1$, minFDE$_6$ and MR$_6$, and attains near-optimal performance in minFDE$_1$ and minADE$_6$, with only marginal differences from the top-performing approaches.
Similarly to some previous works, we further employ the model ensembling strategy with eight models to boost the performance of our model and results are shown in the lower group.
Our proposed HAMF achieves state-of-the-art performance across key metrics with $K=1$ and MR$_6$, while maintaining competitive second-best results in remaining metrics, demonstrating the superior capability in generating accurate and reliable future trajectory predictions.

It is worth noting that our proposed model achieves competitive results with the significantly reduced model complexity and improved inference latency. 
We conduct the efficiency analysis with two state-of-the-art methods QCNet~\cite{zhou2023query} and DeMo~\cite{zhang2024decoupling} which are learnable anchor-based approaches, presented in Table \ref{tab:size}.
We utilize DeMo itself without integrations for the fair comparison.
The evaluations are conducted on AV2 validation set with the batchsize as 1 by one NVIDIA GeForce RTX 3090 GPU.
Experimental results indicate that our model satisfies real-time requirements while maintaining significantly less memory consumption, making it highly suitable for practical autonomous driving applications.
Evaluations on the real-world dataset AV2 indicate that our proposed framework is much more \textit{information effective} for the future motion representation learning, demonstrating state-of-the-art performance with the rather lightweight model architecture. Through the long-range interactions among future motion tokens and scene context tokens in the Attention-based encoder and the efficient modeling with Mamba block during the decoding, accurate and diverse future motion representations are aggregated.

\begin{table}[t]
\caption{Comparisons of model size, inference latency and prediction performance. For all metrics, the lower is the better.}
\vspace{-1.0em}
\tablestyle{8pt}{1.1}
\resizebox{0.47\textwidth}{!}{
\begin{tabular}{x{55}x{40}x{40}x{40}}
\toprule
Methods & Params(M) & Latency & minFDE$_6$  \\ \midrule
QCNet~\cite{zhou2023query} & 7.7 & 94ms & 0.129 \\
DeMo~\cite{zhang2024decoupling} & 5.9 & 38ms & 0.125 \\
\textbf{HAMF (Ours)} & \baseline{\textbf{3.0}} & \baseline{\textbf{22ms}} & \baseline{\textbf{0.123}} \\ 
\bottomrule
\end{tabular}
}
\label{tab:size}
\end{table}

\begin{table}[t]
\caption{Comparisons with two baseline models on AV2 val set.}
\vspace{-1.0em}
\tablestyle{8pt}{1.1}
\centering
\resizebox{0.47\textwidth}{!}{
\begin{tabular}{x{55}x{40}x{40}x{40}}
\toprule
 \centering Models & minADE$_6$ & minFDE$_6$ & MR$_6$ \\ \midrule
 \centering $\mathcal{M}_b$ & 0.667 & 1.333 & 0.164 \\
\centering $\mathcal{M}_q$ & 0.646 & 1.264 & 0.156 \\
\textbf{HAMF (Ours)} & \baseline{\textbf{0.633}} & \baseline{\textbf{1.229}} & \baseline{\textbf{0.145}} \\ 
\bottomrule
\end{tabular}
}
\label{tab:baseline}
\vspace{-2em}
\end{table}

\paragraph{Qualitative results.}
We visualize the qualitative results of HAMF on Argoverse 2 validation set. 
We implement two model variants as baselines for the comparison: the direct-decoding base model $\mathcal{M}_b$ without the usage of learnable future motion tokens and the learnable anchor-based model $\mathcal{M}_q$ that utilize learnable future motion tokens as the queries by cross-attention blocks to capture future motion features with the encoded scene information from the final encoder layer.
Quantitative results are shown in Table \ref{tab:baseline}. 

\begin{figure*}[hbt!]
\begin{center}
\vspace{0.5em}
\includegraphics[width=0.8\linewidth, trim=20 120 100 0]{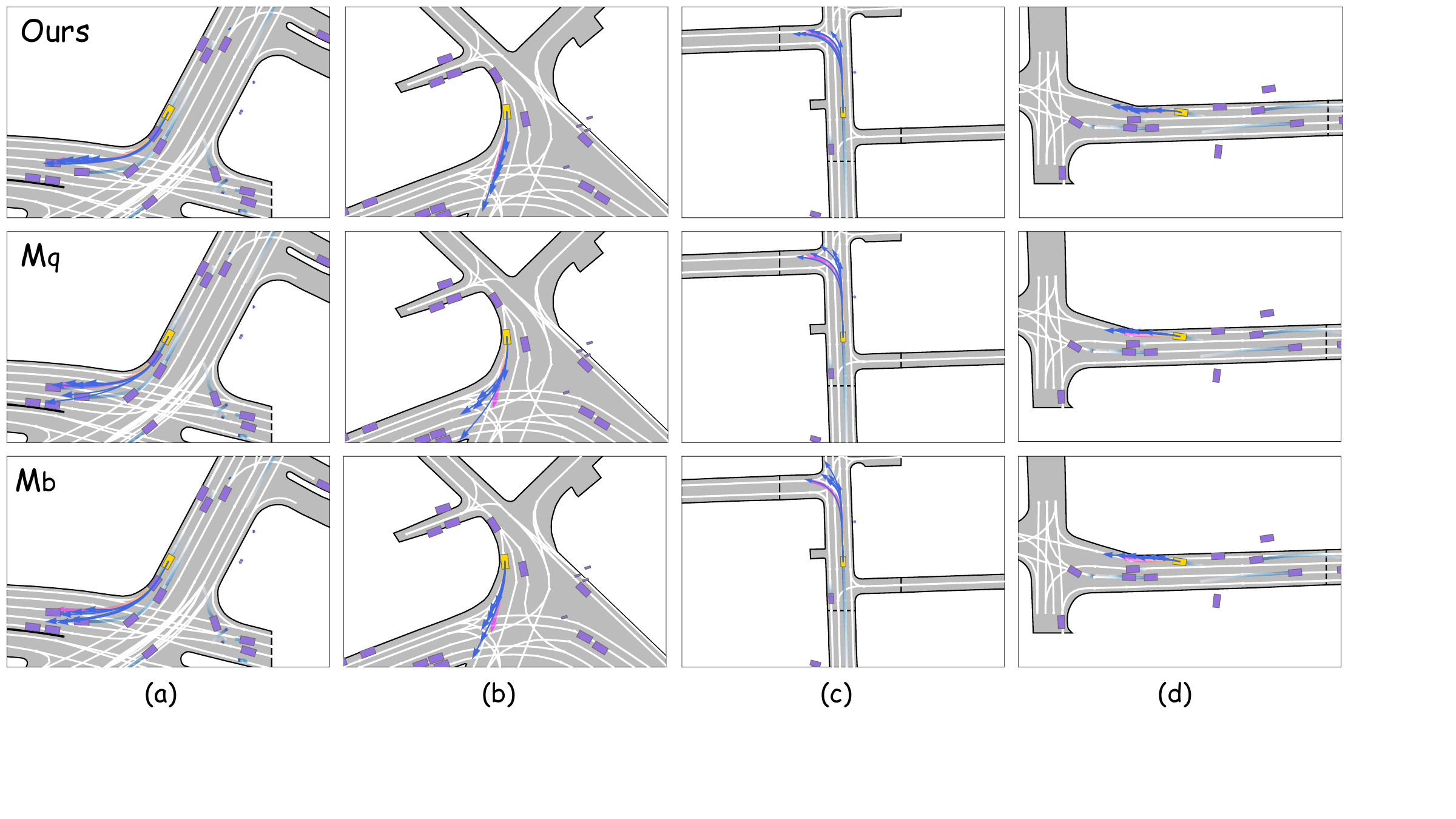}
\end{center}
\caption{Qualitative results of our proposed method, the query-based method $\mathcal{M}_q$ and the base model $\mathcal{M}_b$ on four challenging scenarios of AV2 validation set. Surrounding agents are represented by the bounding boxes in purple and the focal agent in yellow. The line in gradient pink indicates the ground truth and the line in deep blue indicates the multi-modal predicted trajectory.}
\vspace{-1em}
\label{fig:vis}
\end{figure*}

Comparisons of visulation results with the base model $\mathcal{M}_b$ and the query-based model $\mathcal{M}_q$ are presented in Figure \ref{fig:vis}, where results of our HAMF are shown in the first row.
Our model which aggregates future motion features together with scene context encoding predicts more accurate and feasible multi-modal future trajectories, which are closer to the ground truth in the complex driving scenarios (shown in (a), (b) and (d)) and are more compliant to the road layout (shown in (b) and (c)).
Qualitative results demonstrate that our proposed motion forecasting model outperforms the direct-decoding method and the learnable anchor-based method, which obtains the more feasible and comprehensive motion features with the unified scene context understanding and future motion feature learning framework.

\subsection{Ablation Study}
All ablation studies are conducted on Argoverse 2 validation set.
Results of our HAMF are highlighted in the tables. 
To analyze effects of each component within the encoder, we implement three model variants: $\mathcal{M}_1$ that exclusively employs self-attention blocks, $\mathcal{M}_2$ that solely utilizes future motion tokens as queries in the cross-attention blocks and $\mathcal{M}_3$ that utilizes both attention blocks without the interaction in every layer, shown in Table \ref{tab:encoder}. And $\mathcal{M}_b$ denotes the base model without future motion tokens.
Then we compare decoder variants that utilize Bidirectional Mamba~\cite{li2024videomamba, vim} and attention blocks in Table \ref{tab:decoder}, to investigate the effect of our Mamba configuration.  $\mathcal{M}_d$ indicates the model that excludes future motion token modeling in the decoding.
We further study the impact of varying encoder depths and different numbers of future motion tokens, in Table \ref{tab:encoder_depth} and \ref{tab:K}.
Additionally, we implement two variants with distinct encoder architectures: $\mathcal{M}_c$ that utilizes the reversed attention stack in encoder layer and $\mathcal{M}_p$ that employs the parallel attention architecture, with results shown in Table \ref{tab:sequential}.

\begin{table}[t]
\vspace{0.8em}
\caption{Effects of different encoder modules.}
\vspace{-1.0em}
\tablestyle{8pt}{1.1}

\resizebox{0.47\textwidth}{!}{
\begin{tabular}{x{20}x{10}x{10}x{10}x{22}x{22}x{22}}
\toprule
\begin{tabular}[c]{@{}c@{}}Models\end{tabular} &
\begin{tabular}[c]{@{}c@{}}Self \\Attn.\end{tabular} & \begin{tabular}[c]{@{}c@{}} Cross \\Attn.\end{tabular} & \begin{tabular}[c]{@{}c@{}}Inter-\\action\end{tabular} & minADE$_6$ & minFDE$_6$ & MR$_6$ \\
\midrule
 $\mathcal{M}_b$ & &  &   & 0.667 & 1.333 & 0.164 \\
 \midrule
$\mathcal{M}_1$ &\checkmark &  &  & 0.642 & 1.267 & 0.157 \\
$\mathcal{M}_2$ && \checkmark &   & 0.641 & 1.252 & 0.150 \\
$\mathcal{M}_3$ & \checkmark & \checkmark &   & 0.642 & 1.260 & 0.153 \\
\textbf{Ours} &\baseline{\checkmark} & \baseline{\checkmark} &  \baseline{\checkmark} & \baseline{\textbf{0.633}} & \baseline{\textbf{1.229}} & \baseline{\textbf{0.145}} \\
\bottomrule
\end{tabular}
}
\label{tab:encoder}
\end{table}

\begin{table}[t]
\caption {Comparisons of different configurations in the decoder.
} 
\vspace{-1.0em}
\centering
\tablestyle{8pt}{1.1}
\resizebox{0.47\textwidth}{!}{
\begin{tabular}{y{25}x{15}x{22}x{22}x{22}x{22}}
\toprule
\multicolumn{1}{l}{} Methods & Depth & Params(M) & minADE$_6$ & minFDE$_6$ & MR$_6$ 
\\ 
\midrule
\centering $\mathcal{M}_d$ & / & 2.9 & 0.637 & 1.243 & 0.153
\\
\hline
\multirow{2}{*}{Uni-Mamba} & 1 & \baseline{3.0} &\baseline{\textbf{0.633}} & \baseline{\textbf{1.229}} & \baseline{\textbf{0.145}} \\
 & 2 & 3.1 & 0.636 & 1.245 & 0.149 \\
 \hline
\multirow{3}{*}{Bi-Mamba} & 1 & 3.0 & 0.642 & 1.263 & 0.156 \\
 & 2 & 3.2 & 0.635 & 1.241 & 0.152 \\
 & 3 & 3.3 & 0.636 & 1.241 & 0.150 \\ 
 \hline
\multirow{3}{*}{\centering{Attention}} & 1 & 3.1 & 0.639 & 1.252 & 0.153 \\
 & 2 & 3.3 & 0.636 & 1.241 & 0.150 \\
 & 3 & 3.5 & 0.636 & 1.238 & 0.150 \\
\bottomrule
\end{tabular}
}
\label{tab:decoder}
\vspace{-1.5em}
\end{table}

\paragraph{Effects of each component in the encoder.}
We first conduct ablation studies about every component of the encoder, shown in Table \ref{tab:encoder}.
Results of the base model $\mathcal{M}_b$ that is without the usage of learnable future motion tokens are presented in the first row. For other models, we maintain the same decoder architecture with one unidirectional Mamba block and two layers of MLPs to generate the multi-modal trajectories and their probabilities.

With the incorporation of learnable future motion tokens into the self-attention block of each layer, the multi-modal trajectory prediction performance gets improved by 3.75\%, 4.95\% and 4.27\% in terms of minADE$_6$, minFDE$_6$ and MR$_6$.
The performance tends to be better if the future motion tokens are utilized as queries to interact with scene context features in each encoding layer, denoted as $\mathcal{M}_2$.
Then we adopt the cross-attention block in every encoder layer that follows the self-attention block without the interaction mechanism in $\mathcal{M}_3$, that is, the updated future motion tokens from respective attention blocks are summed once at the end of the encoding stage instead of in every layer. Results of $\mathcal{M}_3$ is better than $\mathcal{M}_1$, but worse than $\mathcal{M}_2$, which indicates that the appropriate interaction strategy between two kinds of attention blocks for the feature combination is necessary.
We employ the sum of future motion features in each encoding layer and achieve the best performance with 5.1\%, 7.8\% and 11.6\% improvements in terms of minADE$_6$, minFDE$_6$ and MR$_6$ compared to the base model.

\paragraph{Effects of decoding with Mamba.}
Evaluation results of the model variant $\mathcal{M}_d$ that is without the further modeling of future motion tokens in the decoder are presented in the first row of Table \ref{tab:decoder}. The performance with minFDE$_6$ as 1.243 demonstrates that various motion patterns are preliminarily learned with the effective future motion feature extraction mechanism in the encoder.
To effectively model and capture the complex relationship among multi-model future motion features, we implement the Unidirectional Mamba block, leveraging its proven effectiveness in sequence modeling. 
We conduct the comparison experiments with alternative architectures employing the Bidirectional Mamba and Attention mechanisms with varying depth configurations.

It is shown that our configuration with one block of Uni-Mamba outperforms other methods, with the 5.2\% improvement in MR$_6$ compared to $\mathcal{M}_d$. The performance of model variants with one block of Bi-Mamba and Attention are even worse than $\mathcal{M}_d$. The possible reason is that both Bi-Mamba and Attention perform the bidirectional modeling among the tokens, which potentially disrupts the consistency of the initially learned future motion features.
The experimental results indicate that the more precise modeling requires progressively increasing the number of Bi-Mamba and Attention blocks, which leads to the larger model size with more parameters.
In contrast, the sequential state space modeling mechanism inside Uni-Mamba excels at modeling and maintaining the dependency and coherency within the sequence tokens, producing feasible and distinctive multi-modal future motion representations to enhance the final prediction performance, while over-modeling results in feature degradation as the number of Uni-Mamba blocks increases.

\begin{table}[t]
\vspace{0.8em}
\caption{Results of different encoder depths.}
\vspace{-1.0em}
\tablestyle{8pt}{1.1}
\resizebox{0.47\textwidth}{!}{
\begin{tabular}{x{50}x{25}x{25}x{25}x{25}}
\toprule
Encoder depth & Params(M)&minADE$_6$ & minFDE$_6$ & MR$_6$ \\ \midrule
4 & 2.6 & 0.641 & 1.260 & 0.157 \\
\baseline{5} & \baseline{3.0} &\baseline{0.633} & \baseline{1.229} & \baseline{0.145} \\
6 & 3.4 & 0.629 & 1.226 & 0.145 \\ \bottomrule
\end{tabular}
}
\label{tab:encoder_depth}
\vspace{-2.5em}
\end{table}

\begin{table}[t]
\caption{Results of different numbers of future motion tokens.}
\vspace{-1.0em}
\tablestyle{8pt}{1.1}
\resizebox{0.47\textwidth}{!}{
\begin{tabular}{x{55}x{40}x{40}x{40}}
\toprule
Different $K_e$ & minADE$_6$ & minFDE$_6$ & MR$_6$ \\ \midrule
1 & 0.655 & 1.298 & 0.157 \\
2 & 0.653 & 1.296 & 0.156 \\
3 & 0.643 & 1.263 & 0.155 \\
\baseline{6} & \baseline{\textbf{0.633}} & \baseline{\textbf{1.229}} & \baseline{\textbf{0.145}} 
\\ 
\bottomrule
\end{tabular}
}
\label{tab:K}
\vspace{-2.5em}
\end{table}

\paragraph{Effects of encoder depth.}
Given the proposed unified encoding mechanism, the encoder depth plays a critical role in the prediction performance.
We study the influence of the number of encoder layers in Table \ref{tab:encoder_depth}. 
 Evaluation results demonstrate a 2.3\% improvement in minFDE$_6$, 1.2\% in minADE$_6$ and 5.3\% in MR$_6$ with encoder depth increasing from 4 to 5, while adding one more layer only makes 0.2\% improvement in minFDE$_6$, 0.6\% in minADE$_6$ and no difference in MR$_6$. We finally choose the encoder depth of 5 with the consideration of efficiency.

\paragraph{Effects of the number of future motion tokens.}
The number of learnable future motion tokens impacts the multi-modal prediction feature aggregation substantially. We conduct an ablation study to investigate the effect, shown in Table \ref{tab:K}. For $K_e$ is smaller than 6, we implement the feature projection to get 6 motion modes using the single-layer MLP. Experimental results show a positive correlation between the number of learnable tokens and prediction accuracy, aligning with common sense.

\paragraph{Effects of sequential architecture.} 
We implement a sequential architecture that integrates self-attention and cross-attention blocks within each encoder layer to optimize future motion feature extraction.
For comparative analysis, we evaluate two alternative configurations: $\mathcal{M}_c$ that reverses the attention block by placing cross-attention before self-attention, and $\mathcal{M}_p$, a parallel architecture inspired by~\cite{taskprompter2023} that processes and combines features through separate attention pathways.
As demonstrated in Table \ref{tab:sequential}, our architecture outperforms both alternatives, achieving superior accuracy in capturing future motion representations. 

\begin{table}[t]
\caption{Ablation study on different encoder architectures. 
}
\vspace{-1.0em}
\tablestyle{8pt}{1.1}
\resizebox{0.47\textwidth}{!}{
\begin{tabular}{x{55}x{40}x{40}x{40}}
\toprule
Models & minADE$_6$ & minFDE$_6$ & MR$_6$ \\ \midrule
$\mathcal{M}_p$ & 0.639 & 1.251 & 0.151 \\
$\mathcal{M}_c$ & 0.636 & 1.243 & 0.150 \\
\textbf{HAMF (Ours)} & \baseline{\textbf{0.633}} & \baseline{\textbf{1.229}} & \baseline{\textbf{0.145}} \\ \bottomrule
\end{tabular}
}
\label{tab:sequential}
\vspace{-1.5em}
\end{table}

\section{CONCLUSIONS}

We propose HAMF, a novel motion forecasting framework that jointly learns the future motion representation with the scene context feature extraction, as a new paradigm that unifies scene understanding and future state prediction.
With the unified encoder, comprehensive future motion features are captured from scene features, leveraging both strengths of self-attention and cross-attention mechanisms.
The Mamba block is utilized in the decoding phase to further model the relationship among multi-modal future motion representations to maintain the consistency and dependency of various motion modalities.
Evaluations on challenging Argoverse 2 benchmark demonstrate that our hybrid Attention-Mamba model achieves state-of-the-art results with the lightweight architecture and real-time inference latency, 
which demonstrates considerable potential for implementation in real-world autonomous driving systems.

One of the limitations of our work is the simple feature interaction strategy between self-attention and cross-attention blocks. 
Although accurate future motion representation is obtained by the sum operation, more effective strategies are left to further exploration, such as the incorporation of learning-based feature fusion methods to improve the representation learning from the scene context. 
We leave it for future work.











\bibliographystyle{IEEEtran}
\bibliography{IEEEabrv,ref}

\end{document}